\documentclass{egpubl}
\usepackage{IMET2023}

\WsPaper           

\usepackage[T1]{fontenc}
\usepackage{dfadobe}  

\usepackage{cite}  
\BibtexOrBiblatex
\electronicVersion
\PrintedOrElectronic
\ifpdf \usepackage[pdftex]{graphicx} \pdfcompresslevel=9
\else \usepackage[dvips]{graphicx} \fi

\usepackage{egweblnk} 

\title[Continual RL for Quadruped Robots]%
      {Towards Continual Reinforcement Learning for Quadruped Robots}

\author[G. Minelli \& V. Vassiliades]
{\parbox{\textwidth}{\centering G. Minelli$^{1,2}$
        and V. Vassiliades$^{2}$
        }
        \\
{\parbox{\textwidth}{\centering $^1$University of Bologna\\
         $^2$CYENS Centre of Excellence
       }
}
}

\begin{document}


\maketitle
\begin{abstract}

Quadruped robots have emerged as an evolving technology that currently leverages simulators to develop a robust controller capable of functioning in the real-world without the need for further training.
However, since it is impossible to predict all possible real-world situations, our research explores the possibility of enabling them to continue learning even after their deployment.
To this end, we designed two continual learning scenarios, sequentially training the robot on different environments while simultaneously evaluating its performance across all of them.
Our approach sheds light on the extent of both forward and backward skill transfer, as well as the degree to which the robot might forget previously acquired skills. 
By addressing these factors, we hope to enhance the adaptability and performance of quadruped robots in real-world scenarios.



\end{abstract}  
\section{Introduction}
Quadruped robots are a rapidly developing technology that has become commercially available and increasingly affordable. As demonstrated by the recent \href{https://www.subtchallenge.com/}{DARPA Subterranean Challenge}, these robots possess the potential to autonomously explore subterranean environments, thereby expanding their deployment possibilities across various domains. In addition, they are already finding practical applications in areas such as construction surveying, oil rig maintenance, and search and rescue operations.

Quadruped robots offer immense potential as interactive and semi-autonomous platforms, wherein users provide velocity commands, and the robot responds by adjusting its gait to the best of its abilities. This adaptability would allow the robot to respond to dynamic situations, such as ceasing movement towards a commanded direction upon detecting potential hazards like the edge of a cliff, or navigating around obstacles encountered along the way. The development of controllers to achieve such functionality remains an active area of research.

In recent years, reinforcement learning (RL)-based approaches have emerged as promising methods for developing robust controllers capable of executing dynamic gaits. These methods often necessitate substantial interaction with the environment, leading to an initial utilization in simulation before being transferred to real-world scenarios (known as sim2real robot learning). To address this, NVIDIA introduced Isaac Gym\cite{makoviychuk2021isaac}, a framework that combines physics simulation, and neural network model learning into a single GPU. This integration enables significant acceleration, achieving orders of magnitude speedup compared to CPU-based frameworks. For instance, \cite{rudin2022learning} successfully trained quadruped robots to walk in just 20 minutes, marking a substantial reduction in wall-clock time when compared to CPU-based counterparts.

The approach of Rudin et al. \cite{rudin2022learning} requires the robot trainer to anticipate all possible scenarios the robot may encounter in its lifetime. This approach trains a deep neural network (NN) policy through parallel simulations spanning diverse environments (e.g., flat terrain, sloped terrain, stairs, etc.). Then during deployment the policy is fixed, even if scenarios never seen before are encountered, missing opportunities for improvement.

What if we do extend the training process presenting new challenges to the agent? Would the robot acquire the new skill, while still retaining its previous abilities?
The answer is probably no: if there is absence of mechanisms to retain old information while adapting to new data, NN models may lose previously learned knowledge, a phenomenon known as the catastrophic forgetting problem \cite{mccloskey1989catastrophic}. A typical way to teaching a new skill (scenario) to the robot using such an approach, while retaining its older skills, would be to include the new simulated environment in the list of the previously simulated environments and retrain.




\begin{figure*}
  \centering
  \mbox{} \hfill
  \includegraphics[width=\linewidth]{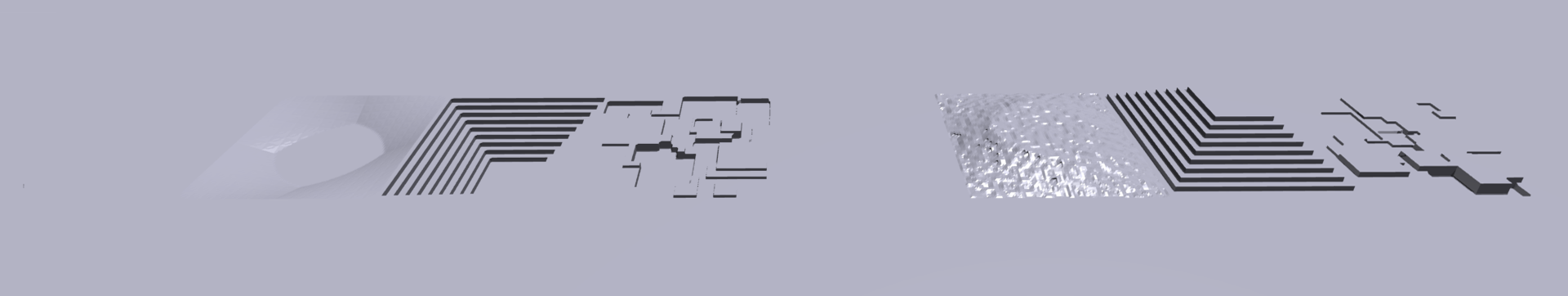}
  \hfill \mbox{}
  \caption{\label{fig:e2h} Easy-to-hard sequence of terrains for robot locomotion training. From left to right: flat, negative slope, stairs down, tiles, flat, positive slope with random roughness, stairs up, tiles. 4096 robots are trained sequentially from easier to harder terrains. After each iteration, 512 robots for each type of terrain are evaluated in parallel.}
\end{figure*}

Our ultimate goal is to develop continual reinforcement learning algorithms \cite{khetarpal2022towards} for robots capable of adding new skills to their repertoire without forgetting and without retraining from scratch.
In this preliminary study, we empirically investigate the performance of the proximal policy optimization (PPO) algorithm\cite{schulman2017proximal}, a state-of-the-art RL method (used in \cite{rudin2022learning}), in the context of environment-incremental learning. Our hypothesis is that by presenting environments sequentially rather than in parallel, PPO will prioritize the current situation at the expense of forgetting previously acquired skills due to its lack of a built-in continual learning mechanism \cite{parisi2019continual, hadsell2020embracing}.


\section{Methodology}

The core idea behind continual learning in our setting is to expose the agent to a structured sequence of environments for a fixed amount of time in order to develop one skill after the other.
Following this approach we are interested in the measurement of the degree of \textit{forward transfer} (i.e., whether learning in one environment bootstraps learning in the environment presented next), \textit{backward transfer} (i.e., whether learning in one environment helps develop better skills for a previously seen environment), and \textit{forgetting} (or negative transfer; i.e., when learning in one environment harms the performance of a previously learned skill).

\subsection{Scenarios}
Building on the work of Rudin et al. \cite{rudin2022learning}, we adopted five types of simulated terrains to test locomotion skills: flat, rough, sloped, stairs, and tiles. The \textit{flat} terrain was considered the easiest, while the \textit{tiles} terrain was viewed as the most challenging. Then, by mixing the order of presentation of these terrains, we created two continual learning scenarios: \textbf{easy-to-hard}, and \textbf{hard-to-easy}. The easy-to-hard scenario (shown in Fig.~\ref{fig:e2h}) introduced the terrains in the following progression: \textit{flat, negative slope, stairs down, tiles, flat, positive slope with random roughness, stairs up, tiles}. The hard-to-easy reverses this order of increasing terrain difficulty.

\subsection{Training and Evaluation}
The training process involves sequentially deploying non-colliding robots on each terrain type. Specifically, the robots start training on one terrain, following random velocity commands, until completion, then are relocated to the next terrain in the predetermined order. For the reward function details we refer the reader to the previous work of Rudin et al.~\cite{rudin2022learning}.

To evaluate the performance of the learned policy in a way that allows observing forward transfer, backward transfer, and forgetting, we establish a validation protocol. During each iteration, we save the learned policy, and in the subsequent iteration, we deploy validation robots with learning switched off. These robots use the policy saved from the previous iteration and operate simultaneously across all terrains alongside the training agents. The performance of the validation robots is assessed by calculating the moving average of the total rewards obtained by the last 100 terminated agents.

\subsection{Experimental setup}
In this setting, we use a modified version of the PPO algorithm that can recognize restarts in the batches. Collecting 24 steps per agent, we split the samples in 4 mini-batches and iterate the learning for 5 epochs. We set a total of 4,000 learning iterations, progressing through the curriculum every 500 iterations. The maximum duration of the agents is set to a constant value of 20 seconds, after which they are restarted to avoid random walkers in the map. Both the actor and the critic are implemented with DNNs composed of three fully connected hidden layers of size 512, 256 and 128 respectively. The networks' input comprises of 48 sensory control measurements, commands, and the previous action, along with terrain height measurements sampled in a square area beneath the agent. These measurements add up to a total of 235 values. The output of the policy network (i.e., the actions) is 12D which corresponds to desired joint positions (the robot's degrees of freedom).

\section{Results}

We conducted experiments using the NVIDIA Isaac Gym simulator with a Unitree A1 quadruped robot model. For each experiment, we collected results from 5 independent runs, executing the simulations on NVIDIA RTX A5000 GPUs. The training phase used 4096 robots in total, with robots dynamically switched between different terrains every 500 iterations. An additional 4096 robots considered for the validation phase were distributed across all terrains, resulting in 512 robots used to validate individual terrain types.

\begin{figure*}[tbp]
  \centering
  \mbox{} \hfill
  \includegraphics[width=0.9\linewidth]{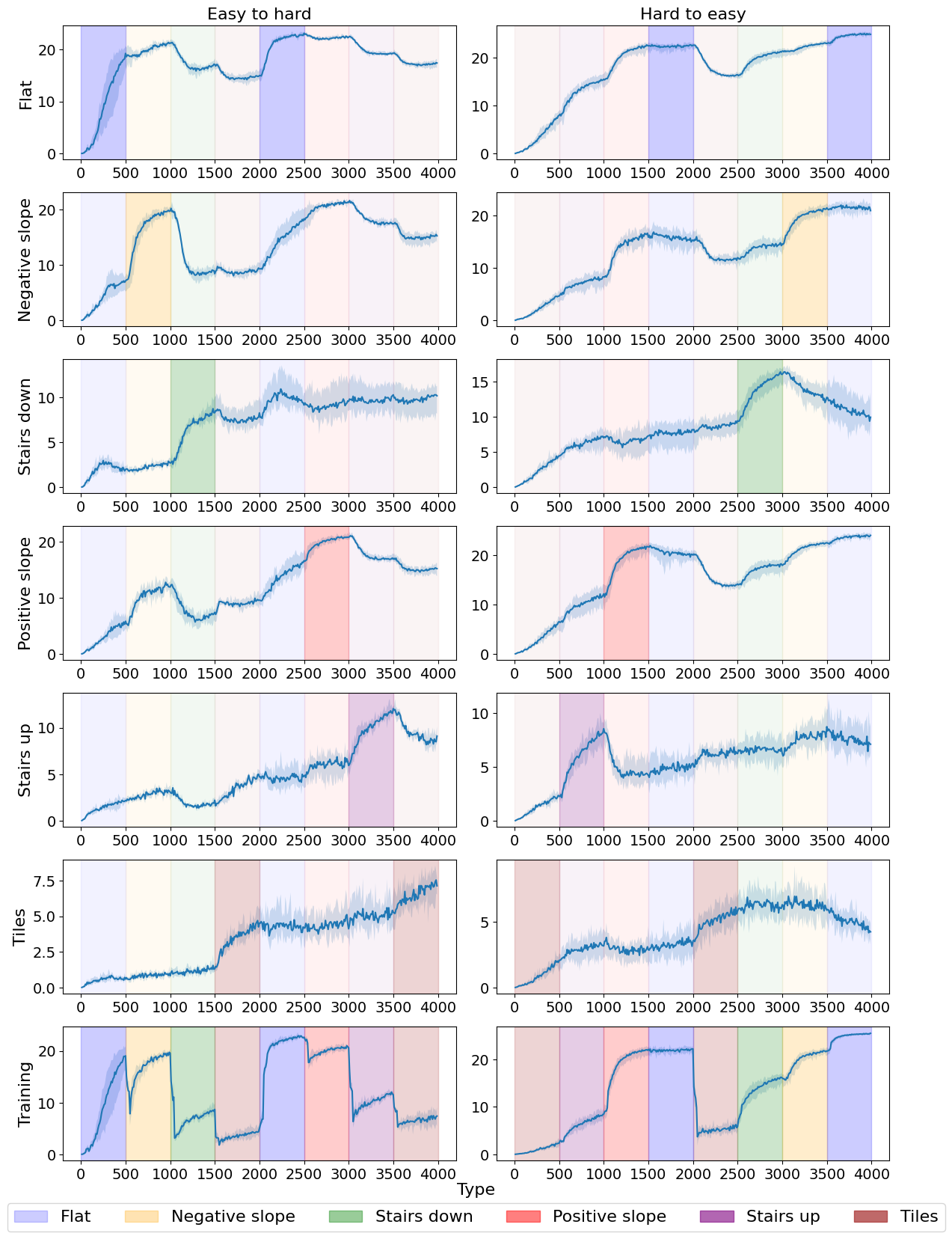}
  \hfill \mbox{}
  \caption{\label{fig:plots} Performance over time of locomotion policies for \textbf{easy-to-hard} (left column) and \textbf{hard-to-easy} (right column) training. Last row shows training, other rows validation on each terrain, with colored bands highlighting when training occurred on that terrain. The blue line represents reward as average over 5 runs, with shaded region as min-max interval. The graphs demonstrate the effects of forgetting and knowledge transfer at different stages of training, the intensity of which is affected by previously acquired skills.
  }
\end{figure*}


\subsection{Easy-to-hard scenario}
As we observe in Fig.~\ref{fig:plots}, in the easy-to-hard scenario, whenever there is a change in terrain during training, the performance of the agent drops and starts increasing again; the exception is when the flat terrain is presented 
again
at iteration 2000, where the performance rises beyond 20 
(performance on flat terrain at 500 iter.). 

During validation, we observe the following. In the initial iterations (0-500) when the agent is trained on the flat terrain, there is an increase in the performance of all skills, more for the sloped terrains and less for the stairs and tiled terrains. When the negative slope is presented (500-1000 iter.), there is some backward transfer, as the skill of walking on flat terrain gets slightly improved; at the same time, this helps improve the skill of walking on a positive slope. When stairs-down is presented (1000-1500 iter.), there is some forgetting of the previous skills as the performance of flat and negative slope decreases; we observe a similar trend for the positive slope and stairs up terrains. 
Training on tiles (1500-2000 iter.) worsens flat and downward stairs walking, improves upward stairs and slope walking, and insignificantly affects negative slope and downward stairs skill.

When flat terrain is presented again (2000-2500 iter.), the skills of walking on negative and positive slopes significantly improve. Positive slope (2500-3000 iter.) also improves the skill of walking on a negative slope and slightly on stairs up. On the other hand, the stairs up terrain (3000-3500 iter.) negatively impacts the flat, positive and negative slope terrain walking skills. Finally, training on tiles for the 2nd time (3500-4000 iter.) negatively impacts the skills in the flat, negative and positive slope, and stairs up terrain, while the skill in the stairs down terrain does not get affected.

\subsection{Hard-to-easy scenario}
In the hard-to-easy scenario, interestingly, the performance of the agent during training gets progressively increased; the exception is when the tiles terrain reappears after the flat terrain. In addition, when the agent trains on the flat terrain for the 1st time (1500-2000 iter.), its performance does not increase, as it is already at a high level. When it trained for the 2nd time (3500-4000 iter.) its performance becomes even better.

During validation we observe the following. The skill on flat terrain generally improves over time with the exception of switching from flat to tiles (2000-2500 iter.) where we observe some forgetting. The negative slope and positive slope walking skills start improving, but when flat terrain is presented (1500-2000 iter.) they start being forgotten, and when the tiles terrain is presented next (2000-2500 iter.) the performance drops even more, which, however, starts increasing afterwards. The stairs down skill improves steadily; at 2500-3000 iterations it increases quickly due to being trained on that terrain, but gets forgotten later as the agents are trained on the negative slope and flat terrains. The stairs up skill gets forgotten when switching to positive slope, but steadily improves later, with the exception of flat terrain at the end (3500-4000 iter.) which negatively (but not significantly) impacts it. 
Finally, the skill of walking on tiles improves slightly with stairs-up training, but declines with negative slope training followed by flat terrain.

\subsection{Discussion}
We observed that learning to move on positive or negative slope generally results in positive transfer. This might be because random velocity commands on a sloped surface might cause the direction to be either uphill or downhill. We did not observe the same when learning to walk up or down the stairs.
The most variance is observed when the agent is trained on the stairs and tiles terrains.


In our observations, we find that the agent exhibits forgetting in both scenarios, which confirms our initial hypothesis. Although the extent of forgetting is lower in the hard-to-easy scenario compared to the easy-to-hard one, it is still evident. To mitigate forgetting, incorporating continual learning mechanisms into the learning algorithm could be a potential solution. Rehearsal, regularization and architectural strategies could be explored. However, implementing these mechanisms is not straightforward with PPO as this is an on-policy algorithm. 
It might be worth considering complementing such mechanisms with off-policy algorithms (e.g., \cite{espeholt2018impala}), or model-based RL (e.g., \cite{wu2023daydreamer}).

\section{Conclusion}

Our research explores the potential adaptability of quadruped robots during deployment. 
We analyzed the effects of exposure to novel situations measuring forgetting and knowledge transfer between skills when agents are trained with the PPO algorithm. 
Adopting two continual learning scenarios, we provided a comparison of the learning dynamics by highlighting how learned skills affect one another at each point in time.
Our results emphasize the importance of developing RL algorithms with explicit continual learning mechanisms in order to enable robots to efficiently adapt to new real-world situations.


\section*{Acknowledgements}
This project has received funding from the European Union’s Horizon 2020 Research and Innovation Programme under Grant Agreement No 739578 and the Government of the Republic of Cyprus through the Deputy Ministry of Research, Innovation and Digital Policy.

\bibliographystyle{eg-alpha-doi}  
\bibliography{refs}        




\end{document}